# Fault detection system for Arabic language

## Riadh BOUSLIM[1], Houda AMRAOUI[2]

[1] University FSJEG Jendouba Tunisia
bouslimi.riadh@hotmail.com

[2] University FSJEG Jendouba Tunisia
houda.amrawi@gmail.com

## 1. Introduction

The study of natural language, especially Arabic, and mechanisms for the implementation of automatic processing is a fascinating field of study, with various potential applications. The importance of tools for natural language processing is materialized by the need to have applications that can effectively treat the vast mass of information available nowadays on electronic forms. Among these tools, mainly driven by the necessity of a fast writing in alignment to the actual daily life speed, our interest is on the writing auditors.

The morphological and syntactic properties of Arabic make it a difficult language to master, and explain the lack in the processing tools for that language. Among these properties, we can mention: the complex structure of the Arabic word, the agglutinative nature, lack of vocalization, the segmentation of the text, the linguistic richness, etc.

In that perspective, our project aims to develop a system to detect errors in spelling, structure and conjugation of the Arabic language. In this article we will proceed as follows. In the first section we'll present some approaches used for the correction of errors. The second section will be devoted to detailed studies of our proposed system. In the last section, we'll perform experimental tests to evaluate the performance of our system.

## 2. State of the art

### 2.1. MASPAR

A multi-agent system is a system of agents' group that communicate with one another to provide answers about a goal to achieve.

MASPAR is a system of analysis of Arabic texts based on the approach of multi-agents. It consists of a set of agents, using a direct communication by sending messages. These agents work together in order to make syntaxes' analysis of a sentence given by the user by determining its syntax composition. (tree, je ne sais pas si ca existe!!!c un mot relativement technique, il faut voir...)

### 2.1.1. MASPAR System Limits

The major drawback of such system is the time taken by the agents for communication and interaction.

One might also note that the MASPAR system does not detect errors of conjugation. Also, it has a non-ergonomic interface.

## 3. Proposed System

### 3.1. General Description

Our system (Figure 1) is designed to detect errors in spelling, structure and conjugation in a non- vowelized Arabic text. It consists of five phases, each uses the information received from the previous phase to finally get a text containing the least number of mistakes.

The segmentation phase consists on dividing the text into sentences and then into words. The lexical phase subsequently receives the word and checks its existence in the database of words.

After verifying that this word belongs to the language, the phase labelling associates the word it has received the possible morph syntactic labels, this makes the word ambiguous, hence the need to remove this ambiguity by passing phase disambiguation, which in applying certain rules, is used to assign to this word the most suitable label.

To correct the word "Wc", we must compare it with the database of words that we have, if this word belongs to our dictionary, it means that "Wc" is a correct word otherwise our system will detect a misspelling.

The algorithm then verify the proper structure of this sentence, otherwise the system will detect a structure fault. Finally, our system is also capable to detect the faults of conjugation in a sentence.

We, first, introduce the general architecture of our system.

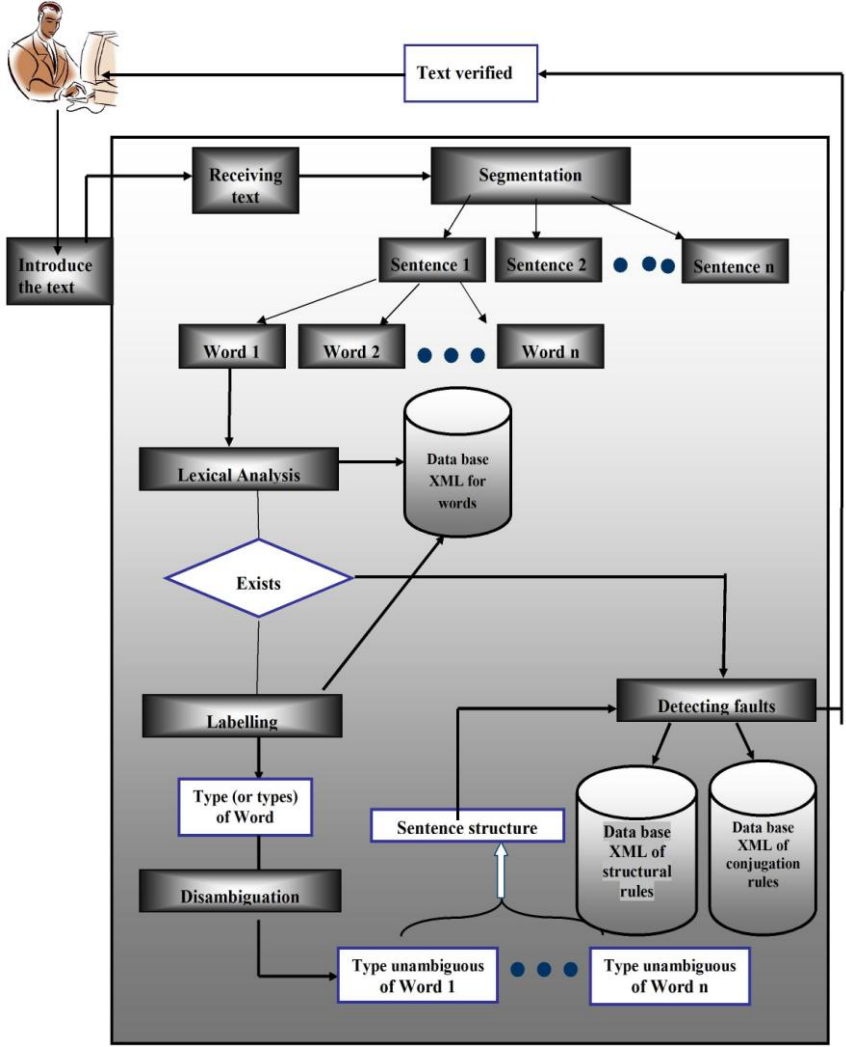

*Figure 1 : Proposed system*

## 3.2. Detailed Description

When receiving an electronic text to analyze, our system launches the first phase which is segmentation. This phase begins with the identification of the text' sentences based on punctuation signs then on the words in each sentence. Subsequently, the words in each sentence will be transferred one by one to the lexical phase.

This will verify whether the word belongs to the language or not by checking its existence in our database of words. Subsequently, the word is sent to the next phase. The phase label is responsible for providing possible morph syntactic characteristics of each received word (from the lexical phase). This means that a word can't go to the labelling phase unless its belonging to our database has been confirmed within the lexical phase.

Because each word can have several labels, the analysis of the word can face certain ambiguity. That's why we must use rules to reduce this ambiguity. Therefore, disambiguation phase is triggered to limit the number of labels associated with the word and assign a single label at a time.

Once the ambiguity is removed, we get into the final phase of the system which role is to apply rules that enable to compare the analyzed structures. This helps detect errors in structure and conjugation.

**ALGORITHM** Editor
**STARTERS:** Wc: the word of the sentence
            Phrase: the input sentence
            BaseXml: the database contains dictionary words
            BaseReglesStruc: the database of structural rules
            BaseReglesConjug: the database according to the rules of conjugation
**START**
      **FOR** each Phrase **DO**
        **FOR** each Wc of Phrase **DO**
          **If**  (Wc, BaseXml) == false **then**
            Write (Wc 'is incorrect')
            **Otherwise**
            Type☐ ReccupererType (Wc, BaseXMl)
               Structurephrase ☐ Type
          **End if**
         **End For**
         Compare (Structurephrase, BaseReglesStruc)

```
          If (compare == true) then
             Write ('the structure of the sentence is not correct')
            Otherwise
            If the structure contains a verb then
                 Apply (BaseReglesConjug, Structurephrase)
            End if
            If (Apply==false) then
               Write ('the combination is not correct')
            End if
       End if
    End For
END.
```

### 3.2.1. Segmentation

This phase consists on dividing the text into sentences and the sentences into words based on markers at the beginning and the end, for example points, semicolons, colons…

### 3.2.2. Lexical Analysis

This phase checks the belonging of each word to the language, obtained from the segmentation phase based on the data base of the words available.

*Verify the existence of the base in the lexicon :* We must ensure that the words introduced constitute the basics of the Arabic language. For that reason, we verify the existence of the base in the lexicon. We have to consult the database of Arabic words, if the extracted base coincides with a word from the database; we conclude that the word exists in Arabic.

### 3.2.3. Labelling

This operation aims to add to the words linguistic information with morphological or syntactic nature in order to identify them.

We have presented several possible tags of the word minimum (prefix + base + suffix):

However, the lack of vocalization does not accurately determine the proper etiquette of the word which causes a certain ambiguity. To reduce this ambiguity, we will proceed to the next step.

*3.2.4. Disambiguation*

A disambiguation is needed to limit the number of labels of these words and subsequently improve the detection of grammatical errors.

*Compatibility Rules:* It can reduce the ambiguity of a word by associating it with one type at a time, so the sentence containing the ambiguous word has more than a structure based on the number of labels that word. Subsequently, the system associates to the word the suitable type according to the structural rules.

*3.2.5. Detecting faults*

For the detection of faults, we can use rules of grammar. These rules describe correct grammatical patterns. For this, we have defined a basic structure rules and another different basis for the conjugation rules. If some text does not match any rule, a structural or conjugal error is detected. To detect structural faults, we'll compare our sentences' structure with the basic structural rules, if this structure does not coincide with any rule, then a lack of structure will be detected, otherwise, if the structure is correct and if it contains a verb, since the combination only applies to the verb, our system will have access to our database conjugation, satisfies a certain compatibility between pre-and post-basic core that typically accompany the verb, if our sentence presents a bad combination when a fault is detected. In the end, the user receives a text containing errors detected with staining of these faults, each depending on the type of errors detected.

*3.2.6.The databases used in the system*

**XML database for the detection of spelling errors**

To detect if the spelling of a given word is correct or not, the verification process run through the XML tree of the dictionary and compare the word with the word list file. It sets out below a portion of our base words.

**XML Data Base for the detection of structural faults**

After ascertaining that the specified words are spelled correctly, we assess at this level whether the sentence structure is coherent or not by comparing it with the base of the structures we have created an XML file with the following form:

```xml
<ReglesApplicables>
  <ReglesPhrasesVerbales>
      <regle>verbe NomPropreFeminin</regle>
      <regle>verbe NomPropreMasculin</regle>
      <regle>verbe NomPluriel </regle>
  </ReglesPhrasesVerbales>
<ReglesPhrasesNominales>
    <regle>NomPropreFeminin verbe </regle>
    <regle>NomPropreMasculin verbe </regle>
```

**XML Data Base for detecting faults conjugation**

Our system can also detect conjugation errors. To handle this, we used an XML file as follows:

```xml
<PronomPersonnel valeur="أنتم">
  <PresentSimple>
      <prebase>ﺗ</prebase>
      <PostBase>ون</PostBase>
  </PresentSimple>
  <PresentNegation>
      <prebase>ﺗ</prebase>
      <PostBase>وا</PostBase>
  </PresentNegation>
</PronomPersonnel>
```

# 4. Test and Validation

We choose to assess the performance criteria that are available: the ergonomics and the response time chosen by our system.

Regarding ergonomics, performance analyzers must have a user-friendly interface, presenting a number of functionality to help users better handle this interface to manage the features offered by the system.

The speed of response is another important constraint for parsers for, to be useful in the real world, they must return a response very quickly.

### 4.1. Experiments

Our experiments on the system relate texts of Arabic literature in various fields. We introduced those relating to the field of Medicine, Marketing, Economics and Arabic grammar.

(-): If no fault is detected.

(+): If an error is detected.

| *Sentences* | *Detection of spelling errors* | *Detection of structural errors* | *Detecting of conjugation errors* |
|---|---|---|---|
| يبحث في أصول تكوين الجملة و قواعد الإعراب | (-) | (-) | (-) |
| و يبحث في في أصول تكوين الجمة و قواعد | (+) | (+) (+) | (-) |
| يتكون جسم الإنسان من أجهزة مختلفة الوظائف | (-) | (-) | (-) |
| التسويق هو مجموعة من العمليات أولأنشطة | (+) | (-) | (-) |
| التسويق هو مجوعة من العمليات أو الأنشطة، تشبعوا رغبات العملاء | (+) | (-) | (+) |
| أنتم لم تذهبون | (-) | (-) | (+) |
| أنتم لم تذهبوا | (-) | (-) | (-) |
| يأخذ إيمان أقراص | (-) | (-) | (+) |
| يأخذ أيمن أقراص | (-) | (-) | (-) |
| تذهبين إيمان | (-) | (-) | (+) |
| تذهب إيمان | (-) | (-) | (-) |
| هما لن يذهبان | (-) | (-) | (+) |
| لم يكتبوا الجملة | (-) | (-) | (+) |
| هم لم يكتبوا الجملة | (-) | (-) | (-) |
| أيمن يذهب | (-) | (-) | (+) |
| النحو العربي هو علم، تبحث في أصول تكوين الجملة و قواعد الإعراب | (-) | (-) | (-) |
| يذكر أن في مثل ذلك المكان | (-) | (+) | (-) |

To evaluate the error detection, we use the rate of accuracy (standard indicator classification [4]). This indicator is between 0 and 1.One being the perfect result.

To calculate this index, we needed to appoint different sets.

Let D be the total set of words, incorrect words D + and D-words correct. D + and D-form a partition of D. Let R be the set of words identified as erroneous. Some words are part of R + D D-other.

The precision is out as an index of the proportion of words identified as erroneous. Its formula is:

Detection Accuracy = | D + ∩ R | / | R |

### 4.2. Results and interpretations:

We note that our system has a good detection for errors in spelling and structure. (Indicator precision = 1 for the detection of spelling errors and 0.75 for structures). Indeed, we get a quick response if the word entered is incorrect or the structure is wrong. We therefore have a very high proportion of errors actually detected. We can also note the good accuracy of the wrong word or structure which facilitates the coloration of errors.

We can also note that our system has a medium detection for errors in conjugation (Indicator accuracy = 0.56). This is for several reasons:

Note first the difficulty of the Arabic language in particular as regards to the conjugation.

 Then, our system took only where the verb is conjugated in the simple present and present Negation and verifies the compatibility between pre-and post-foundation bases with different personal pronouns. By cons, although the average conjugation fault detection, our system provides a new paradigm as it has treated the most difficult        in the Arabic language is the conjugation.

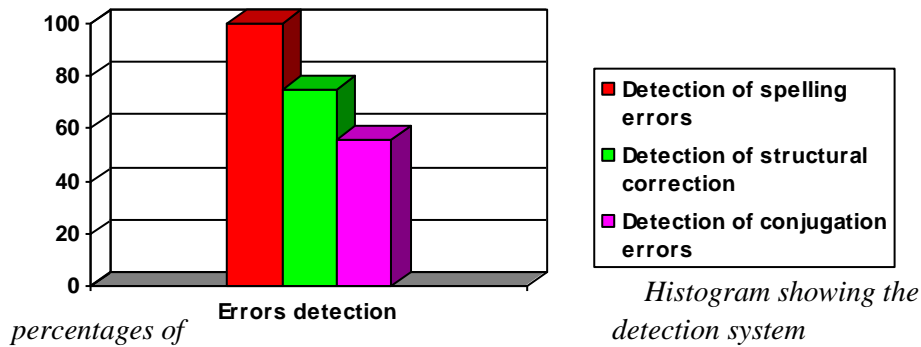

*Histogram showing the detection system*

## 5. Conclusion and Perspectives

The information retrieval and text mining in Arabic is a major challenge. We are interested in this work to develop an application to detect errors in spelling, structure and conjugation in the Arabic text.

The development of this project allowed us to familiarize ourselves with the Java language, a language in the promising field of programming technologies. It allowed us to consolidate our knowledge on various techniques including manipulation of XML.

The work we have done is a response to the objectives set at the outset of the project. However, it can evolve by considering several extension elements.

We can consider adding propositions for the wrong words in order to improve the performance of our system. We can also add more functionality to our tool such as translating the input text from one language to another following the user's choice. We can also handle the case semantics and the texts vowels.

## 6. Bibliographie